\documentclass[runningheads]{llncs}
\usepackage[T1]{fontenc}
\usepackage{amssymb}
\usepackage{amsmath}
\usepackage{hyperref}

\usepackage{booktabs}
\usepackage{tabularx}
\usepackage{ragged2e}
\usepackage{makecell}

\usepackage{graphicx}

\usepackage{listings}
\usepackage{xcolor}

\lstdefinestyle{dto}{
  language=Python,
  basicstyle=\ttfamily\small,
  columns=fullflexible,
  frame=single,
  framerule=0.4pt,
  rulecolor=\color{black},
  backgroundcolor=\color{black!3},
  xleftmargin=0pt,
  xrightmargin=0pt,
  framexleftmargin=-0.1em,
  framexrightmargin=-0.1em,
  aboveskip=0.6em,
  belowskip=0.6em,
  showstringspaces=false,
  tabsize=4
}

\usepackage{enumitem}
\usepackage{xcolor}

\definecolor{workflowgreen}{HTML}{A7B016}
\definecolor{workflowpurple}{HTML}{841F8B}
\usepackage{url}

\begin{document}
\title{X-SYS: A Reference Architecture for Interactive Explanation Systems}
%hi Ho
\author{Tobias Labarta\inst{1}, Nhi Hoang\inst{1},
Maximilian Dreyer\inst{1},
Jim Berend\inst{1}, Oleg Hein \inst{1}, Jackie Ma\inst{1}, Wojciech Samek \inst{1,2,3}, Sebastian Lapuschkin \inst{1,4}}
\authorrunning{Labarta et al.}
\institute{
Fraunhofer Heinrich Hertz Institute, Berlin, Germany \and
Technische Universität Berlin, Germany \and
BIFOLD – Berlin Institute for the Foundations of Learning and Data, Germany \and
Technical University Dublin, Ireland}

\maketitle       % typeset the header of the contribution
\begin{abstract}
The explainable AI (XAI) research community has proposed numerous technical methods, yet deploying explainability as systems remains challenging: Interactive explanation systems require both suitable algorithms and system capabilities that maintain explanation usability across repeated queries, evolving models and data, and governance constraints. We argue that operationalizing XAI requires treating explainability as an information systems problem where user interaction demands induce specific system requirements. We introduce X-SYS, a reference architecture for interactive explanation systems, that guides (X)AI researchers, developers and practitioners in connecting interactive explanation user interfaces (XUI) with system capabilities. X-SYS organizes around four quality attributes named STAR (scalability, traceability, responsiveness, and adaptability), and specifies a five-component decomposition (XUI Services, Explanation Services, Model Services, Data Services, Orchestration and Governance). It maps interaction patterns to system capabilities to decouple user interface evolution from backend computation. We implement X-SYS through SemanticLens, a system for semantic search and activation steering in vision-language models. SemanticLens demonstrates how contract-based service boundaries enable independent evolution, offline/online separation ensures responsiveness, and persistent state management supports traceability. Together, this work provides a reusable blueprint and concrete instantiation for interactive explanation systems supporting end-to-end design under operational constraints.

\end{abstract}
\section{Introduction}
Deep neural networks differ fundamentally from engineered systems: their components emerge from data-driven optimization across billions of parameters rather than explicit design~\cite{amari_backpropagation_1993,goodfellow_qualitatively_2014,fazi_beyond_2021}. What neurons, channels, or attention heads encode remains largely opaque~\cite{guidotti_survey_2018,das_opportunities_2020}, posing risks in safety-critical contexts where understanding failure modes is indispensable~\cite{belle_principles_2021,nikiforidis_enhancing_2025}. Explainable AI (XAI) research has developed numerous techniques, from local attributions~\cite{ribeiro2016should,lundberg2017unified,bach2015pixel}, and counterfactuals~\cite{guidotti2024counterfactual}, to concept-based methods~\cite{kim2018interpretability,achtibat2023attribution,bau2017network}. However, these predominantly address isolated aspects of explainability. XAI studies reveal persistent socio-technical gaps between this algorithmic ``toolbox'' and practitioner requirements for delivering effective explainable experiences~\cite{liao2020questioning,ehsan2023charting}.

Industry reports confirm this gap: organizations recognize explainability as critical, yet fail to operationalize it. A lack of transparency ranks as the second-most reported AI risk but is not among the most commonly mitigated risks~\cite{singla2025state}.

Applying XAI in practice remains challenging. Explainability requirements are shaped by stakeholder needs, application contexts, and user interactions, all changing across the AI lifecycle (e.g., development, validation, monitoring), creating a highly dynamic development environment~\cite{liao2020questioning,bhatt2020explainable,sipos2023identifying,dhanorkar2021needs,langer2021we}. Stakeholder-centered design introduces computational requirements and latency constraints that have to be considered since computational demand between XAI methods can differ by orders of magnitude~\cite{samek2021explaining}. High-risk and regulated environments add further complexity through auditability requirements that mandate reproducible model behavior, user actions, and explanations~\cite{eu2024aiact,raji2020closing,leventi2022deep}.

For example, consider a practitioner using LIME~\cite{ribeiro2016should} to debug a medical model: Waiting for computationally expensive explanations disrupts their workflow (Panel 1). When comparing cases, they cannot trace which model version produced problematic results (Panel 2). Additional XAI methods and workflow requirements cannot be integrated and necessitate rebuilding infrastructure (Panel 3). And when expanding to multiple stakeholders, the system cannot accommodate diverse explanation needs concurrently (Panel 4). These challenges motivate architectural quality attributes to close the deployment gap end-to-end and treat explainability as an information systems problem where stakeholder needs are reflected in interaction design and system capabilities.

We introduce the term \emph{explanation system} (X-SYS) to denote this holistic perspective: systems that generate, orchestrate, and present explanations. Prior work has addressed explanation user interfaces (XUI)~\cite{chromik_human_2021,bove2023investigating,fussl2024explanation}, XAI development processes~\cite{haas2024stakeholder}, and domain-specific architectures~\cite{core2006building,garofalo2023conversational,nwakanma2023explainable,wang2024open}, but a generalizable architecture across explanation and model types remains absent.

Our contributions are twofold:
\begin{enumerate}
  \item A reference architecture for explanation systems (X-SYS) that identifies core components (model services, explanation services, XUI services, data services, orchestration and governance), based on STAR (scalability, traceability, adaptability, responsiveness) quality attributes and specifies their interactions through explicit interface contracts. It also provides system-level design considerations for development and deployment;
  \item An implementation of X-SYS through SemanticLens~\cite{dreyer2025mechanistic}, a system for semantic search and activation steering of component representations in vision-language models, demonstrating how the architecture operationalizes in frontend and backend capabilities.
\end{enumerate}

\begin{figure}[t]
  \centering
  \includegraphics[width=0.9\linewidth]{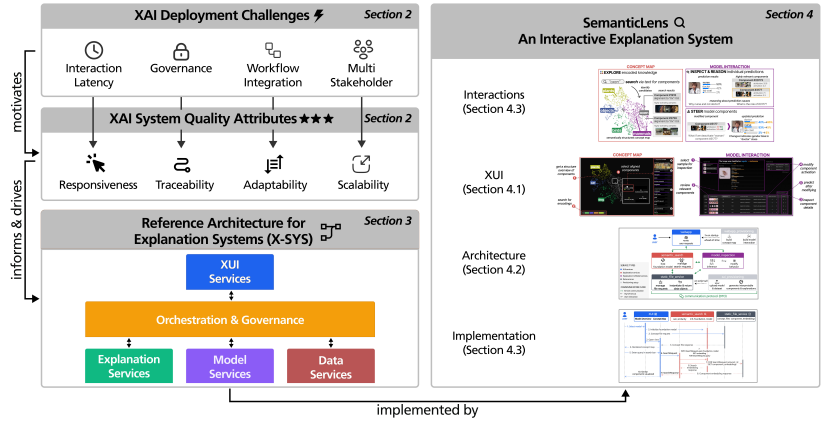}
  \caption{X-SYS Overview: Motivated by XAI deployment challenges, we present quality attributes for XAI systems and informed by it, a reference architecture. SemanticLens, an interactive explanation system, is presented as implementation of the reference architecture.}
  \label{fig:overall}
\end{figure}

Together, these contributions aim to move beyond viewing XAI methods, stakeholder-centered XAI and XUI as isolated aspects and toward treating explainability from a holistic systems perspective. An overview of the contributions and paper structure is presented in \autoref{fig:overall}.

\section{Background}\label{sec:background}
We distinguish explanation user interfaces (XUI) and explanation systems (X-SYS) to avoid conflating two different design problems: (i) how users experience, interpret, and act upon explanations, and (ii) how explanation capabilities are engineered, orchestrated, and maintained. We use \emph{interactive explanation} in an operational sense: an explanation becomes interactive when user actions, in response to XAI, can change which model or explanation is shown, compared, or tested (e.g., switching explanation types, exploring counterfactual changes or comparing prediction instances). This definition focuses on interaction primitives rather than particular explanation methods.

\subsection{Explanation User Interfaces (XUI)} \label{subsec:xui}
Explanation user interfaces constitute the human-facing layer through which explanation outputs are presented, contextualized, and manipulated. Chromik and Butz define XUI as ``the sum of outputs of an XAI system that the user can directly interact with''~\cite{chromik_human_2021}, emphasizing that interfaces are not optional wrappers for explainable models but the primary locus where explainability becomes usable.

The distinction between an explainable model and an explanation interface was formalized in DARPA's XAI program, which frames them as complementary parts of an end-to-end explanation process~\cite{gunning2019darpa}.

Much prior work on XAI provides static XUIs in the form of saliency maps \cite{evans2022explainability,alqaraawi2020evaluating}, example-based rationales~\cite{evans2022explainability,narayanan2024prototype,gallee2024evaluating}, or concept-level summaries~\cite{achtibat2023attribution,kim2018interpretability}. 

XAI does not guarantee user understanding; Effectiveness depends on the ``delivery'', meaning the visual presentation to the user~\cite{lim2023diagrammatization,bo2024incremental}. High attribute counts or complex visualizations can negatively impact cognitive load, while oversimplified explanations can erode trust~\cite{doshi2017towards,abdul2020cogam,nourani2022importance,yin2019understanding}. Explanations therefore have to be delivered at the appropriate level of cognitive demand for effectiveness~\cite{gregor1999explanations,kim2023help,langer2021we}. 

The appropriate level of cognitive demand depends on the stakeholders, which are central to XUI design and development. Relevant groups can include end users, domain experts, policymakers, and management, whose differing goals and constraints shape what an effective XUI must provide~\cite{langer2021we,kim2024stakeholder,vermeire2021choose}.

These differing stakeholder goals imply that XUIs should support different kinds of explanatory activity, often requiring interaction mechanisms that let users steer, question, and refine explanations rather than passively consume a single output.

Shneiderman~\cite{shneiderman2020bridging} distinguishes \textit{Explanatory} and \textit{Exploratory} types of XUI. Explanatory XUIs aim to
convey a single explanation (e.g., a visualization or text). In contrast, exploratory XUIs let users investigate explanations and model behavior through interaction, enabling them to iteratively probe, compare, and refine what the system communicates. Concretely, an interface may combine multiple explanation types in a unified view or offer step-wise complexity calibration via incremental explanations~\cite{bo2024incremental}.

Interactivity or \textit{scrutability} (``allowing users to question and
correct the system'')~\cite{tintarev2007explanations} is an important design consideration for XUI because it supports cumulative learning and the development of users' mental models, which has been associated with successful XAI use~\cite{bansal2019beyond,chromik_human_2021,springer2020progressive,bo2024incremental}. Examples include dialogue-oriented interaction for step-wise sensemaking, as well as control-oriented interaction in which users iteratively modify model or data configurations to steer system behavior toward a desired outcome~\cite{chromik_human_2021}.

These interaction concepts open a potential action space, including drill-down into cases of interest~\cite{tsiakas2024unpacking}; changing the level of explanatory detail and granularity to match users' questions~\cite{liao2020questioning}; switching between visual and textual explanation formats~\cite{tsiakas2024unpacking}; returning to recent interaction context~\cite{amershi2019guidelines}; comparing cases via contrastive outcomes and alternatives~\cite{liao2020questioning,zhu2018explainable}; exploring ``what-if'' variations through counterfactual questions~\cite{liao2020questioning,zhu2018explainable}; and capturing intermediate findings through persistent notes or artifacts~\cite{zhu2018explainable}. They recenter explainability around the user: Beyond XAI outputs, what are users enabled to do, to actively refine their understanding?

Each of these interaction possibilities becomes a system capability requirement. This observation leads to a key implication: interactive XUIs are constrained by what the underlying system can compute, retrieve, version, and expose reliably. Once explanations become interactive, interface design and system architecture become inseparable concerns, because each new interaction primitive requires the support of new backend capabilities.

\subsection{XAI System Development}\label{subsec:xsys}
The main obstacle to operationalizing XAI is no longer absent methods but absent end-to-end development approaches~\cite{bhatt2020explainable,liao2020questioning,langer2021we,vermeire2021choose,haas2024stakeholder}. XAI deployments often prioritize technical stakeholders over end users, with local techniques serving predominantly as internal debugging tools~\cite{bhatt2020explainable}. Research focuses on method development without considering stakeholder needs~\cite{langer2021we}, and practitioners lack guidance implementing the methods~\cite{liao2020questioning,vermeire2021choose}. The deployment gap is fundamentally a system-building gap: explainability goals are rarely transferred into integrated, stakeholder-facing systems~\cite{haas2024stakeholder}.

Recent works propose domain-specific XAI architectures for cloud AI services~\cite{wang2024open}, conversational XAI systems~\cite{garofalo2023conversational}, healthcare~\cite{elbagoury2023hybrid} and industrial domains~\cite{nwakanma2023explainable,aslam2024user}.

To address the broader system-building gap, Haas et al.~\cite{haas2024stakeholder} propose an iterative process model with asynchronous microservices for computational efficiency, providing actionable guidance but limited generalizability on architectural structures.

Adhikari et al.~\cite{adhikari2022towards} propose ASCENT, a standardized ontology that makes XAI solutions describable and searchable across three modules: the AI system, the use case (including user background and task context), and the explanation algorithm. ASCENT provides a common specification layer linking requirements to methods, taking a first step toward planning explainability from a systems perspective. However, it does not address how such requirements should be operationalized in a running system or consider the XUI and interaction perspective.

The deployment gap also shows in current offerings for XAI development: developers get to choose from a variety of post-hoc explanation toolkits (e.g., AIX360~\cite{arya2021ai}, Captum~\cite{kokhlikyan2020captum}, Zennit~\cite{anders2026software}, Zennit-CRP~\cite{achtibat2023attribution}, SHAP~\cite{lundberg2017unified}) attachable to existing model pipelines to generate diagnostic plots or transparency reports. This can be sufficient for one-off inspection, but it becomes challenging once explainability is expected to support recurring workflows, multiple stakeholders, or interactive exploration. The core problem is integration: explanations need to be generated, served, and validated under operational constraints (latency, access control, traceability, versioning), and they must remain aligned with changing models, data, and user goals. 

Interactive explanation systems in particular require responsiveness to maintain analytical flow: Human-computer interaction research establishes that systems must respond within approximately one second to avoid interrupting the users' thought processes~\cite{nielsen1993response}. They require traceability to meet regulatory requirements for audit trails and reproducibility~\cite{goodman2017european,eu2024aiact,leventi2022deep}. To accommodate evolving XAI methods without disrupting operations, the systems should be adaptable~\cite{bass2012software}. And they should also be scalable, to handle varying load from single-user debugging to multi-stakeholder audit scenarios~\cite{dhanorkar2021needs,langer2021we}. 

While the term \emph{XAI system} has been used in recent literature~\cite{lopes2022xai,hachi2024development}, it has historical precedent earlier than much of today's deep-learning-centered discourse. Early work defined XAI systems as those capable of presenting ``an easily understood chain of reasoning from the user's order, through the AI's knowledge and inference, to the resulting behavior''~\cite{van2004explainable}. In this historical line, the military simulation of tactical decision making ``Full Spectrum Command'' (2002) is described as the first known XAI system~\cite{brewster2002using,meske2022explainable}. Notably, Core et al.~\cite{core2006building} presented in 2006 a modular architecture for XAI systems that separated reasoning, logging, and interaction components. However, to our knowledge, XAI systems have not been framed explicitly as an information systems challenge with corresponding architectural considerations for modern ML deployments.   

This matters because both classical system concerns, such as interfaces, access management, and orchestration, and MLOps concerns, such as model serving and data pipelines, directly determine whether explanations remain usable beyond static XUI.

Existing architectural guidance remains limited. \autoref{tab:architecture_comparison} summarizes prior work addressing system-level concerns for XAI. While these approaches provide valuable foundations, they offer partial coverage: Core et al.~\cite{core2006building} addressed component separation but predate modern deep learning systems, ASCENT provides specification without operational guidance and lacks interface considerations, and Haas et al.~\cite{haas2024stakeholder} demonstrate stakeholder-centered development but remain implementation-specific. None address holistically the responsiveness, traceability, adaptability, and scalability required for interactive explanation systems under deployment constraints.

\begin{table}[t]
\caption{Comparison of architectural and system-level guidance approaches for explanation systems. The table evaluates what each approach provides, which architectural concerns it addresses, what capabilities it enables for practitioners, and remaining limitations.}
\centering
\footnotesize
\setlength{\tabcolsep}{4pt}
\renewcommand{\arraystretch}{1.15}

\begin{tabularx}{\textwidth}{@{}
p{0.13\textwidth}
>{\RaggedRight\arraybackslash}X
>{\RaggedRight\arraybackslash}X
>{\RaggedRight\arraybackslash}X
>{\RaggedRight\arraybackslash}X
@{}}
\toprule
\makecell[l]{\textbf{Approach}} &
\makecell[l]{\textbf{Provides}} &
\makecell[l]{\textbf{Addresses}} &
\makecell[l]{\textbf{Enables}} &
\makecell[l]{\textbf{Limitations}}\\
\midrule

Core et al.\ 2006~\cite{core2006building} &
Modular architecture &
Component separation; logging &
Decoupled logic and presentation & 
Dated; limited guidance \\
\addlinespace[0.5em]

ASCENT~\cite{adhikari2022towards} &
XAI solution ontology &
Documentation &
Solution specification and search &
No operational guidance; no architecture \\
\addlinespace[0.5em]

Haas et al. 2024~\cite{haas2024stakeholder} &
Process model with microservice example &
Stakeholder needs; async computation &
Stakeholder-centered workflows &
Implementation-specific; limited abstraction \\
\addlinespace[0.5em]

\textbf{X-SYS (ours)} &
Reference architecture with quality attributes and contracts &
Responsiveness, traceability, adaptability, scalability &
Systematic decomposition and integration &
High-level; requires tailoring \\

\bottomrule
\end{tabularx}
\label{tab:architecture_comparison}
\end{table}

\subsection{From Deployment Gaps to Architectural Requirements}\label{subsec:arch_requirements}
The gaps in \autoref{subsec:xsys} condense to four fundamental challenges for operationalizing XAI systems:

\textbf{Workflow integration challenge}: Practitioners require explanations across multiple model contexts (development, validation, monitoring), explanation methods, and user workflows, yet most XAI toolkits provide static artifacts without persistent state management or consistency across sessions~\cite{dhanorkar2021needs,langer2021we,haas2024stakeholder}. This necessitates \textit{adaptability}: evolving and combining explanation capabilities without interrupting system operations. 

\textbf{Interaction latency challenge}: Interactive workflows require responsive explanation presentation to maintain analytical flow~\cite{chromik_human_2021}, yet computational costs of XAI methods vary by orders of magnitude~\cite{samek2021explaining}, creating tension between explanation thoroughness and user experience. This necessitates \textit{responsiveness}, requiring architectural decisions about computation placement and caching strategies.

\textbf{Multi-stakeholder scalability challenge}: Different stakeholders require different explanation types and workflows~\cite{bhatt2020explainable,dhanorkar2021needs}, from single-user debugging sessions to multi-user audit scenarios. This demands \textit{scalability} in both computational resources and catering to varying stakeholder and use case demand.

\textbf{Governance challenge}: In high-stakes and regulated settings, explanation outputs must be audit-ready. Reconstructing which model, data state, and configuration produced an explanation, and demonstrating compliance during audits. The EU AI Act introduces record-keeping obligations for high-risk AI systems, including automatic logging for system operation traceability~\cite{eu2024aiact}. Algorithmic auditing frameworks emphasize end-to-end documentation linking development decisions, deployment context, and monitoring evidence to observed system behavior~\cite{raji2020closing}. Explanation systems should also control access to model internals and sensitive data. Different stakeholders require different access levels: developers need full access for debugging, while end users receive curated explanations~\cite{bhatt2020explainable}. This necessitates the \textit{traceability} of user actions, roles and system responses.

These challenges translate into four architectural quality attributes that were the drivers of the X-SYS design: scalability, traceability, adaptability, and responsiveness (\textit{STAR}). Together, STAR guide the development of interactive explanation systems to accommodate evolving XAI methods, maintain interactive flow, serve diverse stakeholder and computational demands, and enable auditability and reproducibility.

\section{X-SYS: Towards a Reference Architecture for Interactive Explanation Systems}\label{sec:reference_architecture}

XAI systems face demands that necessitate STAR. Addressing these demands requires an architectural guidance that generalizes beyond single systems while making explicit the structural and behavioral constraints needed to preserve these quality attributes. X-SYS provides this as reference architecture~\cite{cloutier2010concept}, a reusable blueprint on domain level that defines the components, interfaces, and design rules to systematically connect algorithmic capabilities, user-facing design, and system integration.

\subsection{Architectural Components and Responsibilities}
\begin{figure}
 \centering
 \includegraphics[width=0.9\columnwidth]{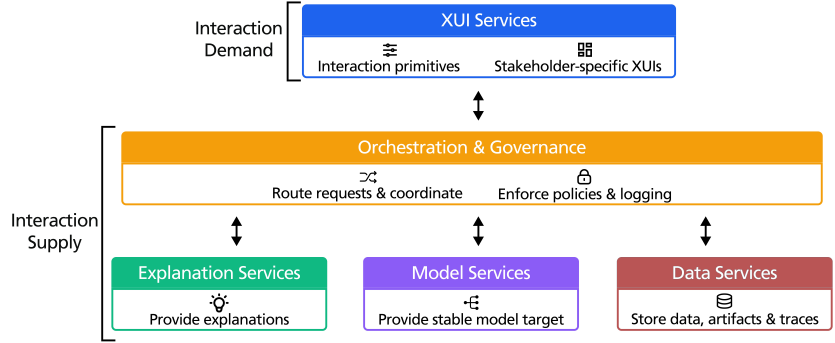}
 \caption{X-SYS reference architecture showing the five core components and their primary interactions. XUI Services request interaction demand that is supplied by Explanation and Model Services. Data Services form the foundational persistence layer, while Orchestration and Governance provide cross-cutting coordination.}
 \label{fig:xsys_architecture}
\end{figure}

X-SYS operationalizes explainability through five components derived from gaps identified in \autoref{subsec:arch_requirements}. The decomposition reflects three concerns: (i) separating user-facing interaction (XUI Services) from computational capabilities (Explanation Services, Model Services), (ii) decoupling persistent state management from transient computation (Data Services), and (iii) cross-cutting coordination and policy enforcement (Orchestration and Governance). Figure~\ref{fig:xsys_architecture} illustrates the component structure and its primary interactions. 

\paragraph{Orchestration and Governance} coordinate service interactions and enforce cross-cutting concerns. Orchestration routes requests between services, manages control flow by selecting execution modes (synchronous for interactive paths, asynchronous for batch operations), coordinates distributed workflows, and implements caching to reduce redundant computation. Governance may enforce authentication and authorization policies, apply rate limiting and maintain audit logs. This component connects all components and thereby enforces adaptability: routing XUI requests to backend services, returning Explanation Service results, querying Data Services for cached results and version information, and monitoring Model Service health and states. Implementation choices include orchestration approaches (API gateways, service meshes, workflow engines, routing layers), interface standardization (e.g., Data Transfer Objects with schema validation~\cite{fowler2012patterns}), and caching approaches (in-memory, distributed caches).

\paragraph{XUI Services} manage user interaction state, render explanation artifacts, and translate user actions into backend service requests. Multiple XUI instances may coexist for different stakeholder groups (model developers, auditors, domain experts) while sharing backend components. XUI Services are decoupled from the non-user facing components via the Governance component. Deployment considerations include interface modality (thin clients via web browser, thick-client desktop applications~\cite{van2023distributed}, programmatic interfaces~\cite{fielding2000architectural}) and state management (client-side via cookies~\cite{barth2011rfc6265} or server-side session management~\cite{calzavara2021measuring}). For XUI design patterns and considerations, we refer to the taxonomy by Chromik and Butz~\cite{chromik_human_2021} and related XUI design work \cite{lei2024focus,calem2024intelligent,jung2025overview,fussl2024explanation}.
 
\paragraph{Explanation Services} compute, compose, and return explanation artifacts for XUI interaction. These services encapsulate diverse explanation methods including feature importance computation~\cite{lundberg2017unified,bach2015pixel}, influential example retrieval~\cite{yolcu2025sparse}, concept probing~\cite{kim2018interpretability,achtibat2023attribution}, and surrogate models~\cite{ribeiro2016should} plus model intervention capabilities like activation steering~\cite{turner2023steering} and prompt engineering~\cite{liu2023pre}. Explanation Services may execute synchronously for interactive queries requiring immediate feedback or asynchronously for computationally expensive batch analyses. Although decoupled through Governance, they require Model Services for inference endpoints, internal activation access, and metadata, while relying on Data Services for reference datasets, cached results, and artifact storage. Implementation choices include method integration approaches (embedded libraries, external service calls, pluggable modules) and data modalities.
 
\paragraph{Model Services} provide stable, versioned access to the predictive models being explained. They serve model predictions through inference endpoints, expose internal representations (activations, embeddings, attention weights) when required by explanation methods, and supply lifecycle metadata including model version identifiers, training data provenance, performance metrics, and drift indicators. Model Services route via Governance requests to Data Services for training datasets, evaluation datasets, and model checkpoint storage and potential Explanation Services for intervention input. Deployment choices for model serving infrastructure and lifecycle management have been thoroughly addressed in MLOps~\cite{kumara2023requirements,raffin2022reference} and ML Systems~\cite{muccini2021software,paakkonen2020extending,wostmann2020conception}. This blueprint treats Model Services as external dependencies and focuses on the interfaces they expose to Explanation Services.
 
\paragraph{Data Services} manage and provide versioned access to all data assets. They store and retrieve input data, reference datasets for XAI methods (e.g., segmentation masks for Network Dissection~\cite{bau2017network}, concept visualizations for TCAV~\cite{kim2018interpretability}), and explanation artifacts with metadata. Critical capabilities include versioning mechanisms for reproducibility (enabling exact reconstruction of prior explanations), provenance and change tracking for auditability (linking artifacts to computational context), and indexed retrieval to support interactive latency. Further capabilities are quality assurance mechanisms, including data validation (schema checking, outlier detection), lineage tracking (linking explanations to specific data versions and quality metrics), and quality metadata calculation. All components request Data Services through the Governance layer: XUI Services retrieve interaction history and user preferences, Explanation Services access reference datasets and cached results, Model Services retrieve training data, and Governance directly requests audit logs and version registries. Deployment variations include implementation (dedicated storage services or embedded within other components), storage technology selection (relational databases, document stores, object storage, time-series databases), indexing strategies, retention policies, and data partitioning schemes. 

\subsection{From Interaction Demand to Interaction Supply}
The architecture distinguishes between \emph{interaction demand} (user actions) and \emph{interaction supply} (system capabilities) (see \autoref{fig:xsys_architecture}). \autoref{tab:xui-xsyst-mapping} maps exemplary interaction patterns to required system capabilities.

\begin{table}
\caption{Some examples for mapping from user actions in XUI (interaction demand) to resulting X-SYS capabilities required to support the demand (interaction supply).}
\centering
\begin{tabular}{p{0.35\linewidth} p{0.65\linewidth}}
\hline
\textbf{Interaction Demand} & \textbf{Interaction Supply} \\
\hline
Change view granularity & Data slicing and aggregation with consistent semantics; indexed retrieval; provenance tracking per view \\
Compare cases or models & Versioned model or data snapshots; stable identifiers; synchronized state across views \\
Explore ``what-if'' &Fast recomputation or approximation; caching; constraint checking; connection of inputs to outputs \\
Switch data modality &Consistent mapping across representations; shared explanation parameters; interoperability across services \\
Return to recent interactions &Interaction logging; state persistence; session management; access-controlled history and replay \\
\hline
\end{tabular}

\label{tab:xui-xsyst-mapping}
\end{table}

These capabilities map to component responsibilities: Data Services provide versioning and stable identifiers, Explanation Services handle fast recomputation with Orchestration caching support, and Data Services maintain interaction logs with Governance access control. Specific realization depends on use case constraints, but the architectural contract remains stable: XUI Services request capabilities through well-defined service interfaces from Governance rather than implementing them directly.

\subsection{Variability Points and Deployment Considerations}
The reference architecture intentionally leaves several aspects open to deployment-specific necessities:

\textbf{Service Granularity:} Components may be deployed as separate services (microservices), combined into larger services (for simpler deployments), or co-located in a single monolith (for prototyping)~\cite{blinowski2022monolithic}. The component boundaries and responsibilities remain consistent across these choices.

\textbf{Communication Mechanisms:} Inter-service communication may use synchronous protocols for interactive paths and asynchronous messaging for batch processing. The choice depends on latency requirements and computational cost.

\textbf{State Management:} XUI state may be managed client-side, server-side, or through distributed session stores depending on scalability and consistency requirements.

\textbf{Explanation Method Integration:} Explanation Services may incorporate methods as embedded libraries, external service calls, or pluggable modules. The architecture remains neutral to integration approach while requiring stable interface contracts.

X-SYS provides a high-level component blueprint as a first step towards conceptualizing and standardizing the development of interactive explanation systems. The following chapter implements X-SYS with the example of SemanticLens.

\section{SemanticLens: Use Case of an Interactive Explanation System}\label{sec:semanticlens}
\begin{figure}
\centering
 \includegraphics[width=1\textwidth]{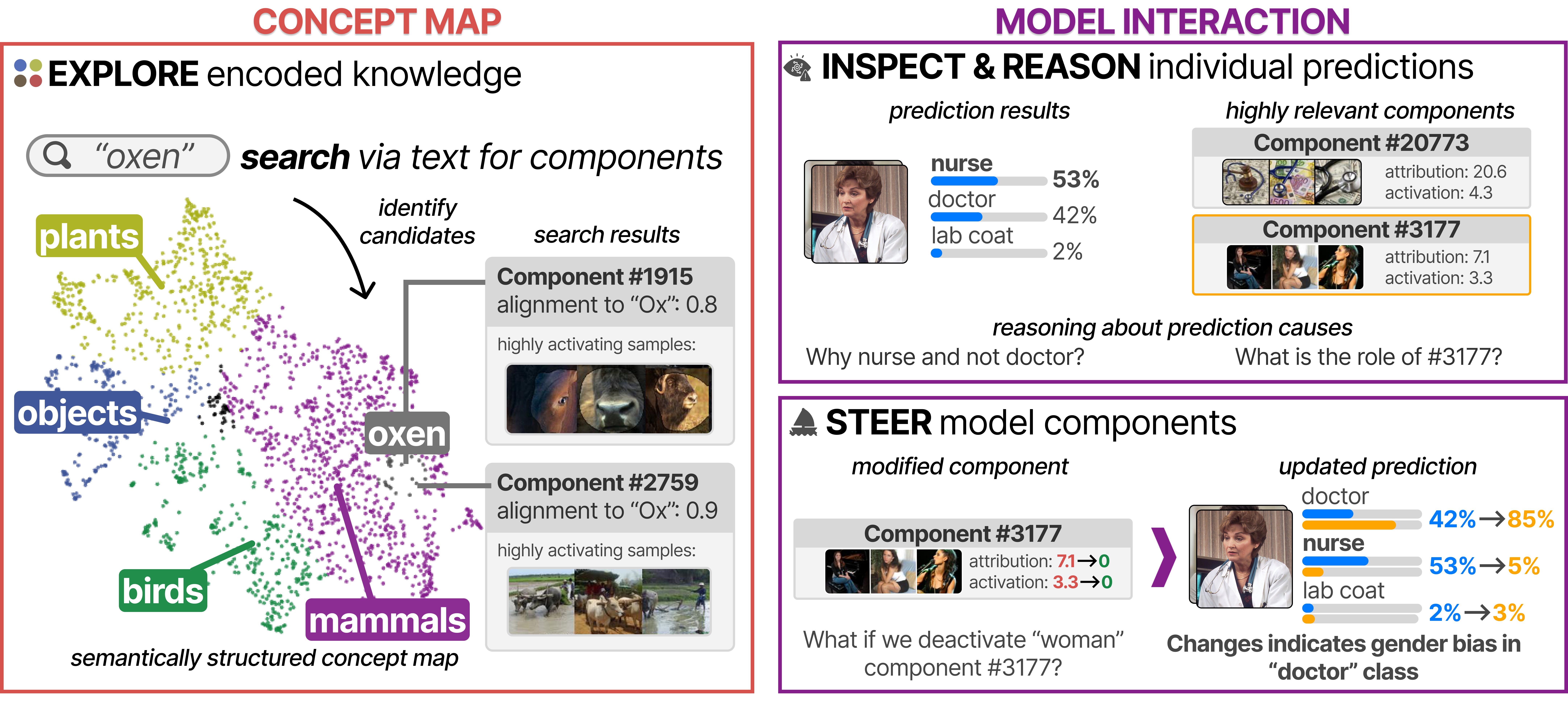}
  \caption{With SemanticLens, users can explore neural representations via text-based search in the \textit{Concept Map} perspective (\emph{left}), or inspect and steer representations at the prediction level in the \textit{Model Interaction} perspective (\emph{right}).}
 \label{fig:overview}
\end{figure}

We present SemanticLens~\cite{dreyer2025mechanistic}, a concrete instantiation that operationalizes X-SYS for concept-based interpretability in vision and vision-language models. SemanticLens addresses three key challenges: (i) making component-level interpretability accessible through natural language queries, (ii) enabling causal investigation through interactive interventions, and (iii) maintaining responsive interaction despite computationally expensive explanation generation (\autoref{fig:overview}).

\subsection{User-Facing Explanation Capabilities}\label{subsec:xui_semanticlens}
SemanticLens provides two complementary perspectives. \textit{Concept Map} enables global exploration: surveying representations, identifying spurious concepts with semantic search, and understanding component-level knowledge organization. \textit{Model Interaction} enables local analysis: inspecting how components drive predictions, testing causal hypotheses through activation interventions, and correcting behavior by suppressing or amplifying components. This bridges between local (observation-level) and global (dataset/model-level) explainability~\cite{guidotti_survey_2018} with a ``glocal'' approach~\cite{achtibat2023attribution}. 

\subsubsection{Concept Map: Global Exploration via Semantic Search.} \label{subsec:conceptmap}
The \textit{Concept Map} provides a global view of model components within a network layer. SemanticLens projects component embeddings into 2D using UMAP~\cite{mcinnes2018umap}, where spatial proximity indicates semantic similarity. Users interact through natural language queries entered via search bar. The system embeds queries using the same foundation model that generated component embeddings, returning components with high cosine similarity. This enables targeted investigation of spurious correlations (e.g., searching for ``watermark'' to identify artifacts). Multiple queries produce blended colors showing semantic overlaps.

\begin{figure}
\centering
 \includegraphics[width=1\textwidth]{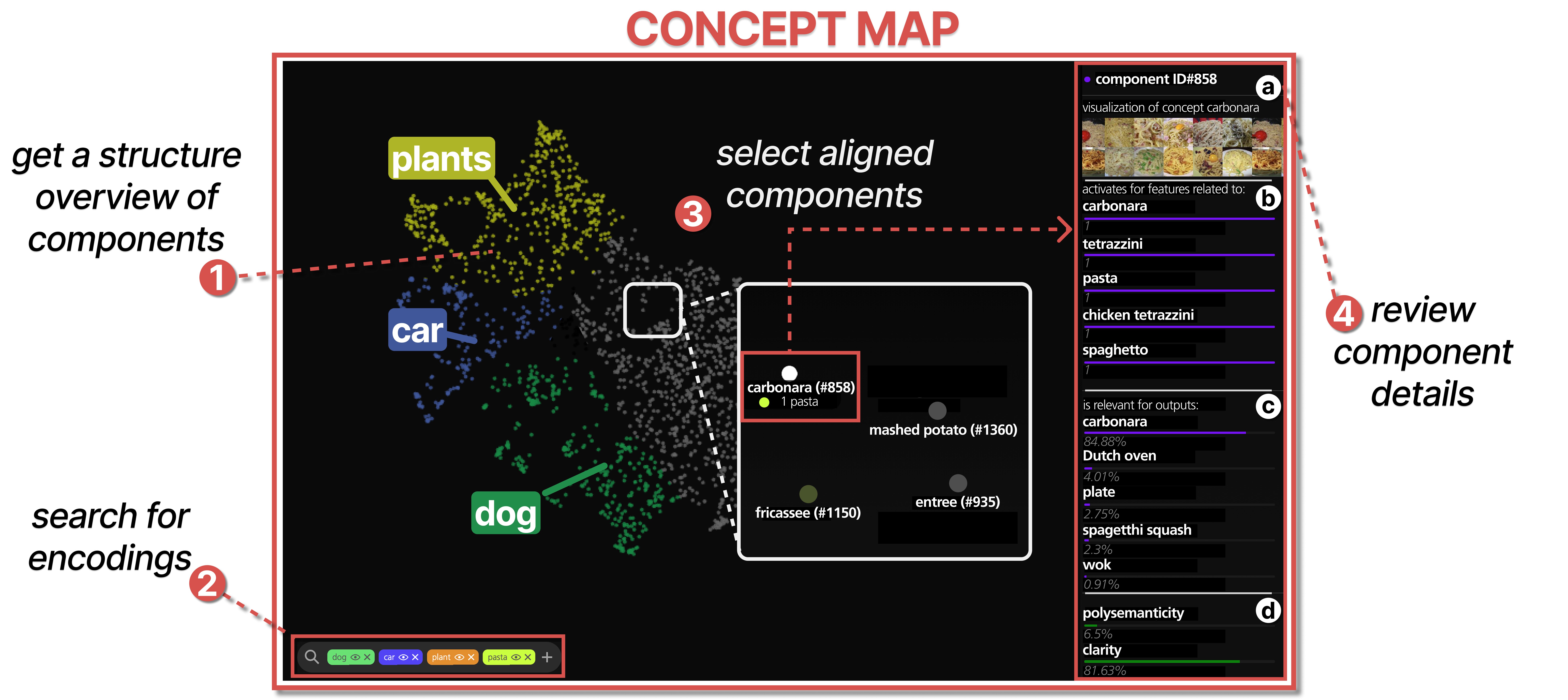}
\caption{The \emph{Concept Map} presents encoded knowledge as clusters of components such as plants, car, dog, food, and their semantic relations. These clusters support users in obtaining an overview of component structure (1). Users can search for encodings (2): the query ``pasta'' in \textit{ResNet50} returns components with high similarity such as ``carbonara (\#858)''. Users can select the aligned components (3) to review their details (4a-d).}
 \label{fig:concept_map}
\end{figure}

Users can select components to inspect detailed properties. As shown in \autoref{fig:concept_map}, these include: Visualization grids of highly activating samples (4a), semantic alignment charts (4b), relevant output classes (4c), and quality metrics (4d). For example, searching ``pasta'' in ResNet50 retrieves component \#858 (``carbonara'', similarity 1.00), plus adjacent components like \#1268 (``dough'', 0.74).

The \textit{Concept Map} enables users to understand what the model has learned globally. To investigate how these learned representations influence specific predictions, SemanticLens provides a complementary local analysis perspective.

\subsubsection{Model Interaction: Local Analysis via Activation Steering.} \label{subsec:modelinteraction}
\begin{figure}
 \centering
 \includegraphics[width=1\textwidth]{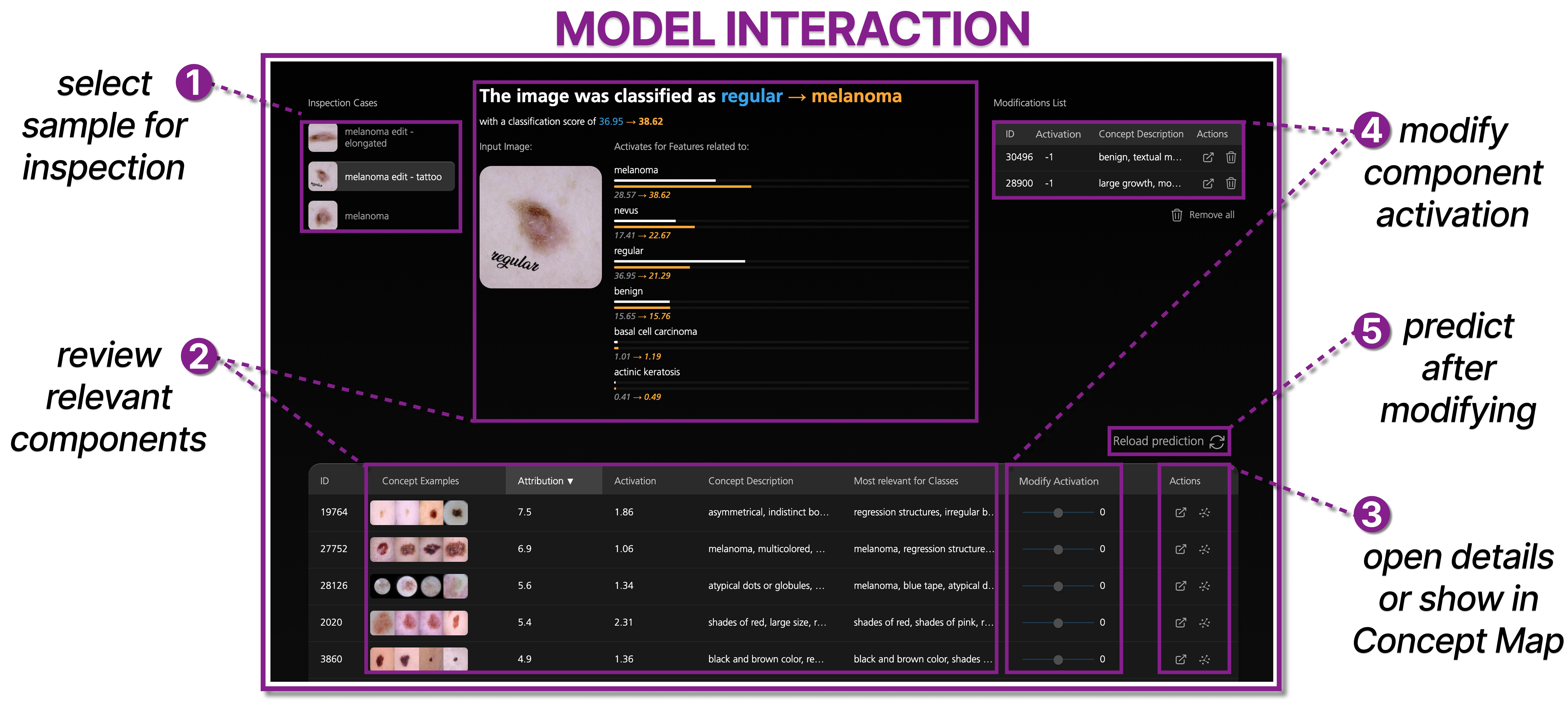}
  \caption{The \emph{Model Interaction} lets users select inspection samples (1), review components (2) and their details (3), and modify component activations (4) to investigate the impact on predictions (5). Yellow text and bars indicate the post-modification state. Shown for the \textit{WhyLesionCLIP} model is a melanoma case, corrected after suppressing the activation of the spurious component \#30496 which responded to textual artifacts (see modifications list).}

 \label{fig:model_details}
\end{figure}

The \textit{Model Interaction} perspective addresses causal influence (\autoref{fig:model_details}). Here, users select input samples, execute inference, and review components ranked by attribution via Concept Relevance Propagation, a method that decomposes model predictions into component contributions~\cite{achtibat2023attribution}. Users perform targeted interventions via activation steering~\cite{turner2023steering}. For neuron $i$ in layer $\ell$ with activation $a_{\ell,i}$, parameter $m_{\ell,i} \in [-1,1]$ rescales as $a'_{\ell,i} = a_{\ell,i}(1+m_{\ell,i})$, enabling hypothesis testing: ``would this prediction change if component \#X responded less strongly?''

A medical model example: Given a melanoma image has added text ``regular'', the model induces misclassification. Component \#30496 showed highest attribution for textual markings. Setting $m=-1$ suppressed this unit, reverting the prediction to ``melanoma'', demonstrating diagnosis and correction of Clever Hans behavior~\cite{lapuschkin_unmasking_2019}.

\subsection{Explanation System Architecture of SemanticLens}\label{subsec:architecture_semanticlens}
\begin{figure}
  \centering
  \includegraphics[width=0.85\textwidth]{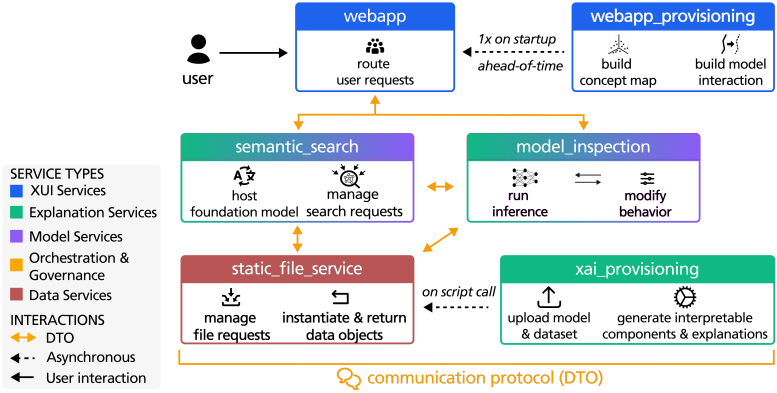}
  \caption{Container architecture of SemanticLens. The legend displays the mapping to the X-SYS reference architecture. At startup, the webapp provisioning service prebuilds the XUI. The webapp routes requests to FastAPI services for semantic search and model inspection, with auxiliary services for static files and explanation provisioning. Components exchange data through defined governance via DTOs (orange arrows).}  
  \label{fig:container_view}
\end{figure}

The SemanticLens architecture satisfies three requirements: (i) responsive interaction, (ii) separation between static artifacts and dynamic queries, and (iii) extensibility across model architectures and XAI methods. \autoref{fig:container_view} shows the architecture and annotates each service with its corresponding X-SYS role (see legend in figure).

A central design decision is separating explanation generation into offline and online phases. Two expensive operations are decoupled: an ahead-of-time webapp build constructs XUI perspectives once on startup, and an asynchronous XAI provisioning service generates interpretable components and visualizations as JSON and images. This offline preprocessing of computationally expensive operations and online serving of interactive queries directly addresses the responsiveness requirement identified in STAR.

Communication follows a contract-based approach using Data Transfer Objects~\cite{fowler2012patterns} that specify data structure and semantics, decoupling XUI from backend implementation and enabling independent service evolution. The two primary backend services (semantic search and model inspection) each encapsulate model access and explanation functionality for their interaction modes. The frontend uses reactive state management for navigating large component spaces. Backend services use Python with FastAPI, relying on Zennit~\cite{anders2026software} for component-wise explanations and MobileCLIP~\cite{vasu2024mobileclip} for semantic search.

This operationalizes X-SYS components: (i) webapp as XUI services, (ii) semantic search and model inspection as model and explanation services, (iii) XAI provisioning and static files as data services, and (iv) DTO protocol as orchestration and governance.

\subsection{Interaction Protocol: Semantic Search Example}\label{subsec:dto_semanticlens}
\begin{figure}
 \centering
 \includegraphics[width=0.85\textwidth]{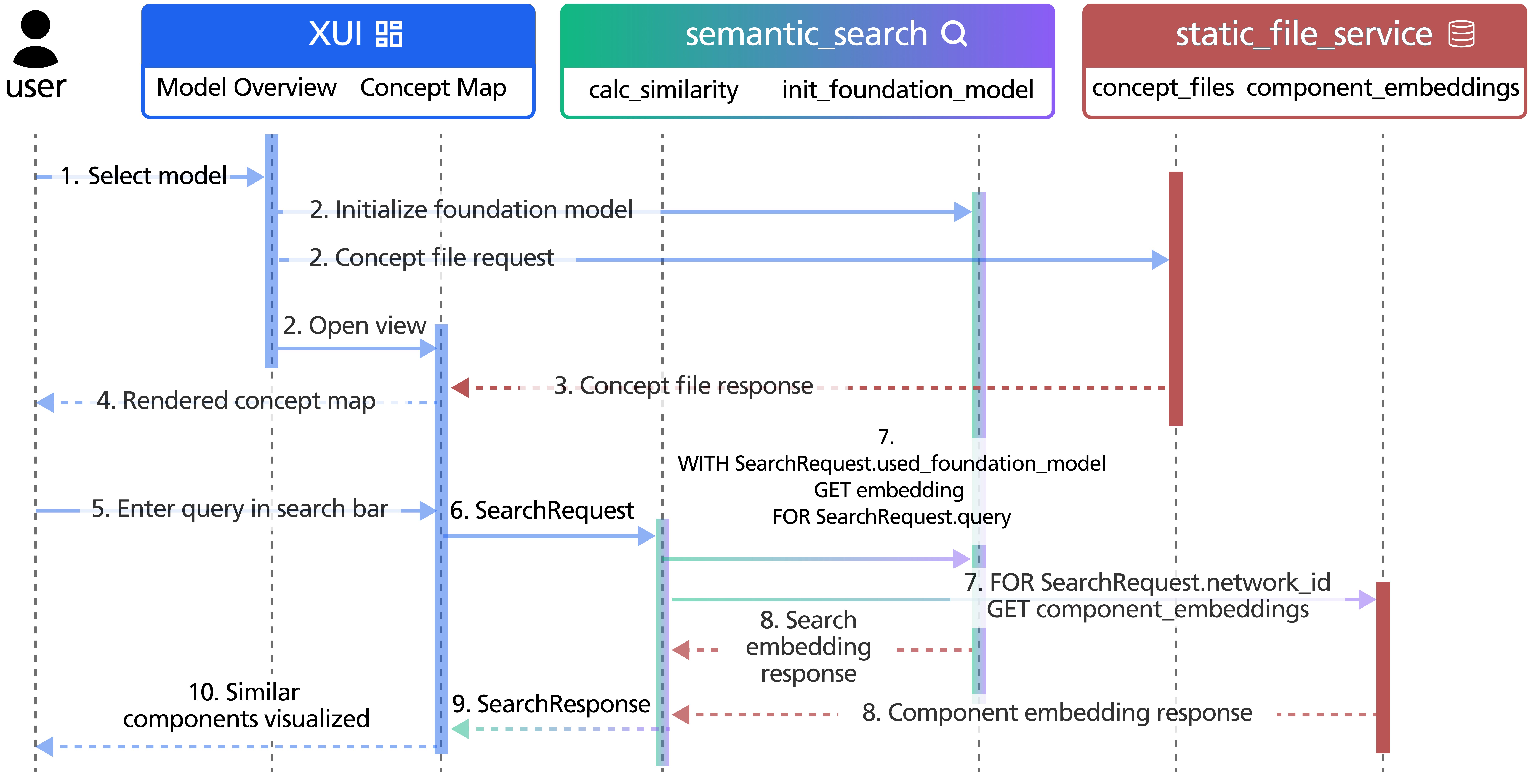}
 \caption{Sequence diagram for semantic search showing interaction between webapp, semantic search service, and static file service. Model selection triggers initialization by loading a foundation model and retrieving precomputed component embeddings. Subsequent search queries are processed through DTO exchanges: The \texttt{SearchRequest} specifies the query demand, the semantic search service requests relevant files from the static file service and computes similarity between query and component embeddings, and the \texttt{SearchResponse} returns ranked results with normalization metadata for consistent visualization.}
 \label{fig:sequence}
\end{figure}
To illustrate how user interactions map to system capabilities, we trace a semantic search request through SemanticLens via the sequence diagram (\mbox{\autoref{fig:sequence}}). This demonstrates how DTO exchanges coordinate distributed services and bridge interaction demand and supply.

Model selection in the Model Overview triggers initialization: loading the foundation model for semantic search and retrieving precomputed Concept Map assets from the static file service. These preparation steps (steps 1-4 in \autoref{fig:sequence}) ensure subsequent queries execute against a fixed model context and reduce runtime cost by executing computation-heavy steps once on startup.

After rendering the Concept Map, users enter search queries captured as \texttt{SearchRequest} DTOs at the service boundary (steps 5-6). The DTO encapsulates retrieval parameters binding the request to model context and search configuration. The semantic search service obtains a query embedding from the foundation model and requests component embeddings from the static file service (steps 7-8). After receiving both responses, the service computes semantic alignment and instantiates the \texttt{SearchResponse} DTO.

\begin{center}\begin{minipage}{0.85\linewidth}
\begin{lstlisting}[style=dto, caption={DTO interface Python code for semantic search. \texttt{SearchRequest} encapsulates query parameters: the search query list, network identifier linking to model context, and foundation model specification. \texttt{NeuronAlignment} pairs component identifiers with alignment scores. \texttt{SearchResponse} returns ranked neuron alignments with normalization metadata (\texttt{min\_alignment}, \texttt{max\_alignment}) enabling the \textit{Concept Map} to map scores to visual properties for consistent rendering.}, captionpos=t, label={lst:dto_search}]
from pydantic import BaseModel

class SearchRequest(BaseModel):
  query: list[str] | None = None
  network_id: str
  used_foundation_model: str # e.g., "clip_whylesion"

class NeuronAlignment(BaseModel):
  neuron_id: int
  alignment_score: float

class SearchResponse(BaseModel):
  query: str
  neurons: list[NeuronAlignment]
  max_alignment: float
  min_alignment: float
\end{lstlisting}
\end{minipage}
\end{center}

The interface contract (\autoref{lst:dto_search}) demonstrates X-SYS principles in practice. \texttt{SearchRequest} explicitly declares what the XUI needs to communicate (user query, model context, search configuration), establishing a stable demand specification. \texttt{SearchResponse} defines the supply structure: ranked component alignments with normalization metadata. The alignment fields provide ranges for consistent visual encoding, enabling the \textit{Concept Map} to map alignment scores to visual properties (color opacity, blending) while maintaining reproducibility across sessions. This protocol-driven design realizes traceability (binding interactions to system states), adaptability (services can evolve independently behind stable interfaces), responsiveness (separation of offline and online computation), and scalability (standardized interfaces enable horizontal scaling of individual services under load).

\section{Conclusion}\label{sec:conclusion}
Operational XAI requires moving beyond isolated explanation methods toward interactive systems shaped by stakeholder workflows and deployment constraints. We introduce X-SYS, a reference architecture framing interactive explanation as a system design problem where XUI interaction patterns induce capability requirements organized around four quality attributes (responsiveness, traceability, adaptability, scalability) that drive architectural decisions. Deployment constraints, including responsiveness requirements and audit-ready traceability across evolving model and data states, are treated as first-class design drivers instead of downstream implementation details. X-SYS specifies a five-component decomposition, maps interaction patterns to system capabilities, and emphasizes stable interface contracts decoupling XUI evolution from backend operations. We instantiate X-SYS through SemanticLens, demonstrating how semantic search, component-level attribution, and intervention workflows are realized via explicit service boundaries and DTO-based interfaces.

Several aspects have not yet been considered. First, the architecture is demonstrated through a single implementation. Generality across domains and modalities and against other architectures remains unvalidated. Also, its actual benefit for XAI system designers remains unvalidated. Second, X-SYS was derived from selected prior work and design-driven reasoning informed by SemanticLens development rather than systematic literature review. The resulting abstraction remains high-level to allow for tailoring, but it potentially limits guidance for concrete engineering concerns (security, privacy, governance, reliability engineering). Third, systematic measurements for the four quality attributes under realistic load and latency budgets have not been proposed. 

These limitations motivate several next steps. First, validation across multiple case studies (non-vision domains, diverse stakeholder contexts) to test whether X-SYS components and dependencies remain stable across implementation variations. Second, systematic mapping of XAI systems and XUI literature to extract recurring capability requirements and architectural patterns, refining X-SYS into an evidence-based reference architecture with explicit coverage claims.
Third, defining measurable benchmark suites that combine existing XAI evaluation~\cite{hedstrom2023quantus} with system benchmarking (workload models, latency and throughput targets, reproducibility checks, traceability audits) to make system attributes empirically assessable. Fourth, standardizing interface contracts and schema evolution strategies to enable composable, reusable explanation capabilities across systems without locking implementations to specific XUIs or toolkits. Fifth, tighter integration with system governance (access control, audit logging, lifecycle alignment with model and data versioning) to provide deployable engineering guidance.
Finally, validating the impact of interactive explanation systems on user understanding and mental models would provide empirical insight about the value of this architectural complexity.
\section*{Acknowledgements}
This work was supported by the Federal Ministry of Research, Technology and Space (BMFTR) as grant BIFOLD (01IS18025A, 01IS180371I); the German Research Foundation (DFG) as research unit DeSBi [KI-FOR 5363] (459422098); and the European Union’s Horizon Europe research and innovation programme (EU Horizon Europe) as grant TEMA (101093003).

\section*{Declaration on the Use of Generative AI}
During the preparation of this work, the main author used Claude Sonnet 4.5 and Grammarly for spelling and grammar checks, paraphrasing and rewording. After using these tools, all authors reviewed and edited the content as needed and take full responsibility for the publication's content.

\bibliographystyle{splncs04}
\bibliography{library}

\end{document}